\documentclass{article}

\usepackage{PRIMEarxiv}

\usepackage[utf8]{inputenc} 
\usepackage[T1]{fontenc}    
\usepackage{hyperref}       
\usepackage{url}            
\usepackage{booktabs}       
\usepackage{amsfonts}       
\usepackage{nicefrac}       
\usepackage{microtype}      
\usepackage{lipsum}
\usepackage{fancyhdr}       
\usepackage{graphicx}       
\graphicspath{{media/}}     
\usepackage{xcolor}
\definecolor{myreferencecolor}{RGB}{0,0,255} 
\hypersetup{
    colorlinks=true,
    linkcolor=myreferencecolor,
    citecolor=myreferencecolor,
    urlcolor=myreferencecolor
}
\title{Comprehensive Overview of Named Entity Recognition: Models, Domain-Specific Applications and Challenges}

\author{
  Kalyani Pakhale \\
 Innover Digital \\
  \texttt{kalyani.pakhale@innoverdigital.com} \\
}

\begin{document}
\maketitle

\begin{abstract}
 In the domain of Natural Language Processing (NLP), Named Entity Recognition (NER) stands out as a pivotal mechanism for extracting structured insights from unstructured text. This manuscript offers an exhaustive exploration into the evolving landscape of NER methodologies, blending foundational principles with contemporary AI advancements. Beginning with the rudimentary concepts of NER, the study spans a spectrum of techniques from traditional rule-based strategies to the contemporary marvels of transformer architectures, particularly highlighting integrations such as BERT with LSTM and CNN. The narrative accentuates domain-specific NER models, tailored for intricate areas like finance, legal, and healthcare, emphasizing their specialized adaptability. Additionally, the research delves into cutting-edge paradigms including reinforcement learning, innovative constructs like E-NER, and the interplay of Optical Character Recognition (OCR) in augmenting NER capabilities. Grounding its insights in practical realms, the paper sheds light on the indispensable role of NER in sectors like finance and biomedicine, addressing the unique challenges they present. The conclusion outlines open challenges and avenues, marking this work as a comprehensive guide for those delving into NER research and applications.
\end{abstract}


\section{Introduction}
Named Entity Recognition (NER) stands as a cornerstone task in Natural Language Processing (NLP), possessing immense significance in information extraction and knowledge organization. NER involves the identification and classification of named entities, such as names of individuals, organizations, locations, dates, and more, within textual data. Its pivotal role lies in its capacity to disentangle structured information from unstructured text, thereby enhancing data retrieval, analysis, and understanding. This research paper embarks on a comprehensive survey of AI techniques applied to NER, encompassing a spectrum of methodologies ranging from conventional to cutting-edge. It elucidates the underlying principles of NER, shedding light on the intricate interplay between linguistic patterns, contextual cues, and machine learning algorithms that enable the recognition of named entities. NER's utility extends across various domains, including information retrieval, question answering{\cite{devlin2018bert}}, document summarization, and language understanding. It serves as the foundation upon which complex NLP applications are built.

The most important point of this research endeavor revolves around the systematic examination of AI techniques for NER. The survey commences{\cite{li2020survey}} with classical rule-based approaches{\cite{palshikar2013techniques}}, where predefined rules guide the identification of named entities. It subsequently explores sequence-based methods that leverage the power of contextual information and sequential patterns for entity recognition. The paradigm shift ushered in by transformer models, epitomized by the groundbreaking BERT{\cite{devlin2018bert}} architecture, has revolutionized NER. This paper delves into the transformative impact of transformers{\cite{devlin2018bert}}, highlighting their contextual embeddings and the innovative fusion of BERT with Long Short-Term Memory (LSTM){\cite{chiu2016named}} and Convolutional Neural Networks (CNN){\cite{chiu2016named}} architectures. These techniques harness contextual embeddings to unravel the complexities of named entity recognition, underscoring the potency of modern NLP architectures.

The study further extends into domain-specific NER models, tailored to the specialized challenges of particular domains. Models such as ViBERTgrid{\cite{lin2021vibertgrid}}, meticulously designed for finance and legal documents, and BioBERT{\cite{lee2020biobert}}, fine-tuned for the intricate landscape of medical NER, exemplify the adaptability of NER techniques to context-specific requirements. Beyond traditional approaches, this research paper delves into the realm of reinforcement learning{\cite{permatasari2021combination}}, unveiling the potential of Gaussian prior{\cite{yang2023gaussian}} and Distantly Supervised NER techniques{\cite{ding2022distantly}}. Deep learning initiatives, including E-NER{\cite{zhang2023ner}} with its evidential learning paradigm, as well as large language model (LLM){\cite{thoppilan2022lamda, brown2020language,chowdhery2022palm}} fine-tuned models and zero-shot mechanisms{\cite{brown2020language}}, showcase the adaptability and prospects of these approaches. Additionally, the paper explores Optical Character Recognition (OCR){\cite{bhatia2019comprehend}} techniques, introducing the role of OCR in the NER pipeline. Prominent solutions from industry leaders such as AWS Textract, Azure, and Google AI demonstrate the synergy between OCR and NER, enriching information extraction capabilities.

As the survey progresses, it culminates in the practical application of NER techniques across diverse domains. The paper illuminates the significance of NER in financial applications, emphasizing its role in parsing financial texts and documents. It concurrently delves into the critical role of NER in biomedical applications, where extracting knowledge from medical texts is of paramount importance. Challenges intrinsic to these domains are dissected, providing insights into the complexities of NER in practice. Lastly, the research paper addresses open challenges awaiting exploration in NER. These challenges transcend domain boundaries and serve as an invitation to researchers and practitioners to venture into uncharted territories, unlocking new possibilities and avenues for advancement. Hence, this comprehensive survey of AI techniques for NER encapsulates the evolution, intricacies, and practical implications of a foundational task within NLP. It caters to the burgeoning interests of researchers and practitioners, elucidating the ever-evolving landscape of NER and its indispensable role in the world of language understanding and information extraction.

\section{Background}
\subsection{What is NER?}

Named Entity Recognition (NER){\cite{Anandika2019ASO, palshikar2013techniques, lee2020biobert}} is a part of text processing used in tasks like extracting information and making sense of text in the Semantic Web. It's about spotting different types of named things in a bunch of documents. These things can be common like names of people, places, dates, and even web links or phone numbers. More advanced NER, called fine-grained NER {\cite{hofer2023minanto, duvsek2023improving}}, digs deeper into categories like specific kinds of people or things, like actors, athletes, or musicians within the larger "PERSON" group. There's also NER for special fields{\cite{radford2018improving,duvsek2023improving}} like biology, where it finds things like proteins and genes, and in manufacturing, where it identifies products and brands.

Nested Named Entity Recognition (NER){\cite{palshikar2013techniques}} is a step ahead of regular NER{\cite{li2020survey}}. It deals with entities that have a hierarchy or parts inside them, like how "Google India" contains both "Google" and "India." This is useful for different languages and fields like finance, medicine, and law, where documents have complex formatting and need special methods to understand entity relationships and boundary positions.

\subsection{Overview: AI Methods for NER} 

Named Entity Recognition (NER) {\cite{li2020survey}} methods can be categorized into three primary groups. It's important to note that while this classification aids in understanding the overarching technique categories, many NER systems discussed in the literature{\cite{li2020survey, palshikar2013techniques}} employ a combination of these approaches. The first category is Rule-based Approaches{\cite{Anandika2019ASO, li2020survey}} where experts manually devise rules to identify specific types of named entities. These rules draw from syntactic, linguistic, and domain-specific knowledge {\cite{Anandika2019ASO}}. The second category, Supervised Learning Approaches {\cite{li2020survey}}, involves creating a sizable manually tagged dataset, where human experts explicitly label instances of the designated entity type. Machine learning algorithms then generalize from this labeled training data to derive NER rules{\cite{duvsek2023improving}}. The third category, Unsupervised Approaches {\cite{radford2018improving}}, typically involves equipping the system with a limited set of seed instances (e.g., city names like 'New York', 'Boston', 'London', 'Seoul'). The system analyzes the provided document collection and learns rules from sentences containing these entity instances. These rules are subsequently applied to detect new instances of named entities, leading to the development of fresh rules. The learning process continues iteratively until no further rules can be uncovered.

Modern deep learning techniques {\cite{Anandika2019ASO, li2020survey}} for Named Entity Recognition (NER){\cite{Anandika2019ASO}} encompass a spectrum of models including recurrent neural networks (RNNs), convolutional neural networks (CNNs){\cite{lample2016neural}}, Long Short-Term Memory (LSTM){\cite{lample2016neural}}, and transformers[27] such as BERT{\cite{devlin2018bert}} and GPT{\cite{brown2020language}}. These techniques hold substantial significance due to their inherent capacity to capture intricate contextual relationships{\cite{peters2018dissecting}} and semantic features within text data. RNNs{\cite{Anandika2019ASO}} and LSTMs{\cite{lample2016neural}} excel in sequence modeling, capturing dependencies between words. CNNs{\cite{chiu2016named}}, on the other hand, can effectively capture local patterns and are particularly useful for character-level representations. Transformers{\cite{alberti2019bert, lin2022pretrained}} exemplified by BERT{\cite{alberti2019bert, devlin2018bert}}, and GPT{\cite{brown2020language}}, revolutionize NER with their attention mechanisms, enabling the model to consider all words in a sentence simultaneously. BERT{\cite{devlin2018bert, novo2023explaining, alberti2019bert}} for instance, offers contextualized embeddings{\cite{peters2018dissecting}}, enriching the understanding of words within their broader linguistic context. The significance of these techniques{\cite{Anandika2019ASO}} lies in their ability to leverage massive amounts of unlabeled data through pre-training, which enhances their performance{\cite{han2023information}} on NER{\cite{Anandika2019ASO}} tasks with limited labeled data. Moreover, the fine-tuning{\cite{hofer2023minanto, han2023information, chowdhery2022palm}} of these models on domain-specific{\cite{duvsek2023improving}} data further tailors their effectiveness{\cite{radford2018improving}}, thus contributing to state-of-the-art NER accuracy and advancing the field of natural language processing.

\section{Methodology}

The below questions provide a broad overview of the concepts, techniques, challenges, and applications related to Named Entity Recognition. Researchers and practitioners often explore these questions to deepen their understanding and improve NER systems.

1) What are the most effective techniques and models for Named Entity Recognition across various domains, and how can they be adapted to address domain-specific challenges?

2) What are the practical applications of NER in finance and biomedical domains, and what insights can be gained from real-world implementations?

\subsection{BERT}
BERT{\cite{devlin2018bert}}, is a Bidirectional Encoder representation from Transformers, that differentiates itself from recent models for language representation like those introduced by Peters et al.{\cite{peters2018dissecting}} and Radford et al.{\cite{radford2018improving}} BERT is uniquely designed to perform deep bidirectional representation pre-training using unannotated text{\cite{li2020survey}}. This is achieved by concurrently considering both the left and right context across all layers. Consequently, the pre-trained BERT{\cite{devlin2018bert}} model can be fine-tuned with the addition of just one output layer, resulting in the creation of cutting-edge models applicable to a wide spectrum of tasks, including Named Entity Recognition{\cite{novo2023explaining}} and language inference. Notably, these enhancements are achieved without significant alterations to the task-specific architecture. BERT's conceptual simplicity is matched by its empirical potency. It attains new state-of-the-art outcomes across eleven diverse natural language processing benchmarks.

  \begin{figure}[h]
  \centering
  \includegraphics[width=0.6\textwidth]{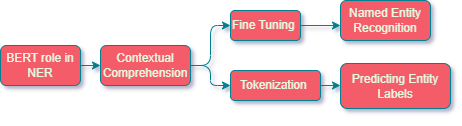}
  \caption{Architecture Overview of BERT: A Bidirectional Encoder Representation from Transformers for NER}
  \label{fig:BERT}
\end{figure}

In the context of Named Entity Recognition (NER), BERT{\cite{alberti2019bert}}, a Bidirectional Encoder representation from Transformers[4], offers a powerful approach. By fine-tuning{\cite{alberti2019bert}} the pre-trained[28] BERT model with a specific NER task, the model can learn to understand and identify named entities within the text{\cite{lin2022pretrained}}. The process involves providing the pre-trained BERT{\cite{lee2020biobert}} model with annotated NER data, where entities are labeled. The model then learns to associate contextual information around words with their corresponding entity types. During fine-tuning{\cite{lin2022pretrained}}, only the output layer of the BERT{\cite{novo2023explaining}} model needs to be adjusted, making it efficient for adapting to NER tasks. Once fine-tuned, the BERT{\cite{alberti2019bert}} model can accurately predict named entities in various documents, even in scenarios where entities have complex relationships and contextual dependencies. This capability makes BERT a valuable tool for achieving high-quality named entity recognition across different languages and domains {\cite{lin2022pretrained}}.

The challenges{\cite{lin2022pretrained}} in improving the detection of semantically similar but non-identical news articles {\cite{novo2023explaining}}, and websites such detection is pivotal for search engines and recommender systems to eliminate duplicates and combat overspecialization{\cite{Anandika2019ASO}}. {\cite{novo2023explaining}} uses fine-tuned BERT{\cite{devlin2018bert, alberti2019bert}} with the SHAP library insights into the model's decision-making process and the significance of individual words and named entities in making accurate determinations{\cite{Anandika2019ASO}}.

\subsubsection{ViBERTgrid}

The key-information Extraction task involves the extraction of entities from various invoices, receipts, and bills.  The deep learning\cite{li2020survey}, neural networks\cite{lample2016neural}. ViBERTgrid\cite{lin2021vibertgrid}, a multi-modal document representation technique that fuses BERTgrid with Convolutional Neural Networks (CNNs)\cite{chiu2016named} which collectively capture textual, layout, and visual information, improving overall representation\cite{baviskar2021multi} ability over BERT\cite{alberti2019bert} contextual embeddings. strengthening of BERT contextual comprehension and CNN training together, ViBERTgrid \cite{lin2021vibertgrid} gains the ability to interpret state-of-the-art techniques and bridge the gap between image segmentation and traditional NLP approaches in document analysis. It reinforces the notion that combining diverse data sources leads to richer representations and improved results. As a result, ViBERTgrid \cite{lin2021vibertgrid} offers a promising avenue for addressing real-world document understanding challenges in various domains, including finance, legal, and data processing. In comparison to competitive methods like LayoutLM, LayoutLMv2, PICK, TRIE, and VIES, our ViBERTgrid demonstrates competitive performance. Even with multimodal domain-specific pretraining on millions of document images, ViBERTgrid achieves comparable results to LayoutLMv2-Base when using BERT-Base and RoBERTa-Base\cite{liu2019roberta} as text embedding models. LayoutLMv2-Large, while slightly more accurate, has significantly more parameters than ViBERTgrid (RoBERTa-Base).

\subsubsection{BioBERT}
BioBERT\cite{lee2020biobert}, is a pre-trained language representation model specifically designed for entity recognition, relation extraction, terminology normalization, and other relevant tasks in the field, It is a Pre-training on large-scale corpora from different domains to learn general language representations, which can then be fine-tuned on medical tasks. It investigates how the pre-trained language model, BERT\cite{devlin2018bert} can be adapted for biomedical corpora and describes concepts like, self-attention, transformer architecture, and the mechanisms behind contextual word, embeddings. By adapting the pre-trained model to the specific task by training additional task-specific layers while keeping the pre-trained weights fixed or partially updating them such as fine-tuning methodology, such as learning rates, optimizer choices, and batch sizes. Standard metrics such as precision, recall, F1 score, or task-specific metrics has been used while, \cite{lee2020biobert} It highlights how pre-training on large-scale biomedical corpora enables the models to learn general language representations that can be fine-tuned for specific biomedical tasks, leading to improved performance and efficiency. It provides insights into the comparative and highlighting advantages and limitations of BioBERT\cite{lee2020biobert} in various biomedical text-mining tasks. It enables researchers to process and extract valuable information from biomedical literature more efficiently, potentially accelerating discoveries and advancements in the biomedical field. It could explore avenues for future research, such as fine-tuning strategies, incorporating additional domain-specific knowledge, or exploring new biomedical text-mining tasks. PromtNER\cite{ashok2023promptner,zhang2023promptner} is combined with BERT zero-shot\cite{brown2020language} and KNN clustering algorithms where a group of entities has been clustered. For example, "Google", "Apple", and "Amazon" combine in one cluster for an Organization entity.

\subsection{Reinforcement Learning}

Reinforcement Learning (RL) is a facet of machine learning focusing on how software agents optimize cumulative rewards through actions\cite{li2020survey}. Agents learn from interactions and rewards
viewing an environment as a stochastic finite state machine.
This framework encompasses the environment's state transition, observation, and reward function. The overarching goal of Reinforcement Learning (RL) is for agents to acquire effective state-update functions and policies for maximizing cumulative rewards.

\subsubsection{Gaussian Prior Reinforcement Learning}

The Gaussian distribution is introduced by the need to address drawbacks in existing approaches to solving nested Named Entity Recognition (NER) tasks. The sequence-based, span-based, and hypergraph-based methods such as label proliferation, exposure bias, computational costs, and ignoring dependencies between nested entities. To overcome these challenges, the research\cite{yang2023gaussian} explores structural and semantic characteristics of nested entities, including entity triplet recognition order and boundary distance distribution that allow the model to learn an optimal recognition order and remove the constraint of predefined triplet orders and introduces\cite{yang2023gaussian} the concept of utilizing Gaussian distribution to better capture the patterns and relationships more effectively.

The Gaussian prior reinforcement learning model\cite{yang2023gaussian} for nested NER consists of 3 interconnected components. The Entity Triplet Generator (ETG) by employs a pre-trained seq2seq model to predict entity boundaries and types in a sentence, with an Index2Token mode converting predicted indices into practical tokens. The Gaussian Prior Adjustment (GPA) incorporates a Gaussian distribution to enhance entity boundary recognition by adjusting the boundary distribution based on distances between tokens. The Entity Order Reinforcement Learning (EORL) component optimizes the model's performance by utilizing Reinforcement Learning, enabling the generation of entity triplets without a fixed order while rewarding high-quality triplets. These components collectively contribute to an integrated framework for improved nested NER by enhancing entity recognition, boundary adjustment, and recognition order learning.

The effectiveness of modules in the GPRL\cite{yang2023gaussian}  model is demonstrated through ablation experiments. Removing the Gaussian Prior Adjustment (GPA) component results in decreased F1 scores on ACE 2004, ACE 2005, and GENIA datasets, highlighting GPA's role in enhancing recognition of nested entity boundaries. Deleting the Entity Order Reinforcement Learning (EORL) module leads to further F1 score reduction, indicating the importance of reinforcement learning in learning proper entity recognition order and mitigating training inference gaps. GPA\cite{yang2023gaussian} is shown to improve attention to neighboring tokens and diminish the impact of distant ones for nested boundary identification. EORL effectively addresses labeling and error identification issues, while also capturing dependencies and interactions between nested entities, leading to enhanced nested named entity recognition and classification. The proposed GPRL\cite{yang2023gaussian} model achieves state-of-the-art performance on nested NER tasks, leveraging Gaussian prior adjustment and reinforcement learning mechanisms.

\subsubsection{Distantly Supervised NER with RL}

The significance of named entity recognition (NER) for extracting valuable insights from digital libraries to advance scholarly knowledge discovery. However, the scarcity of annotated NER datasets, particularly in scientific literature beyond the medical domain, hinders the utilization of advanced deep-learning models. To address this, the study\cite{yang2018distantly} explores distant supervision as an alternative to generate annotated datasets automatically from external resources, despite introducing noise. The focus is on noisy-labeled NER under distant supervision, with a novel Category-oriented confidence calibration (Coca) strategy proposed to account for varying confidence levels towards different entity categories. It integrated into a teacher-student self-training framework, demonstrates promising performance against advanced baseline models, offering a versatile solution for enhancing NER accuracy and easily integrating with other confidence-based model frameworks.

The \cite{yang2018distantly} presents an innovative approach to enhancing Named Entity Recognition (NER) by synergistically integrating Distant Supervision, Partial Annotation Learning, and Reinforcement Learning (RL) methodologies. It addresses the challenges posed by limited annotated data and maximize the NER model's performance. By combining these Reinforcement Learning techniques\cite{yang2023gaussian}, contributes to advancing the field of NER by demonstrating improved results in NER tasks compared to existing methods.

\subsection{Deep Learning}

\subsubsection{E-NER}

E-NER\cite{zhang2023ner}, a specialized Named Entity Recognition (NER) model designed for legal texts, has emerged as a transformative tool for enhancing legal information extraction and understanding. Legal documents are characterized by their complex language, unique terminology, and the presence of numerous legal entities and references. E-NER addresses these challenges by providing a dedicated solution for identifying and classifying legal entities such as statutes, cases, regulations, and legal citations.
  \begin{figure}[h]
  \centering
  \includegraphics[width=0.6\textwidth]{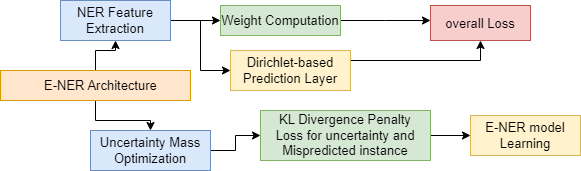}
  \caption{Architecture of the E-NER Framework with Evidential Deep Learning Model}
  \label{fig:NER}
\end{figure}

 The architecture\cite{zhang2023ner} works on two uncertainty-guided loss terms and training strategies to handle sparse and Out-of-Vocabulary/Out-of-Domain (OOV/OOD) entities. E-NER adapts to diverse NER paradigms and showcases enhanced performance, including improved OOV/OOD detection and better generalization for OOV entities, making it a promising approach for robust and reliable NER.
 
 E-NER\cite{zhang2023ner} framework introduces two uncertainty-guided loss terms in addition to conventional Evidential deep learning, accompanied by a range of uncertainty-guided training strategies. The experiments were conducted on 3 paradigms namely sequence labeling, span-based, and Seq2Seq demonstrates that E-NER can be effectively applied to various NER\cite{Anandika2019ASO} paradigms, leading to accurate uncertainty estimation. Furthermore, in comparison to state-of-the-art baselines, the proposed E-NER framework achieves improved OOV/OOD detection performance and enhanced generalization capability on OOV entities. The E-NER\cite{zhang2023ner} with Evidential deep learning model aims to enhance the reliability of NER systems in open environments by introducing a trustworthy NER framework (E-NER) that effectively addresses challenges related to uncertainty and entity recognition.
\subsection{Fine-tuned Large Language Models}
In the rapidly advancing field of NLP, significant attention has been directed toward the large language models(LLM) such as GPT-3\cite{brown2020language}, LaMDA\cite{thoppilan2022lamda}, and PaLM\cite{chowdhery2022palm} which achieve remarkable results in zero-shot and few-shot\cite{fritzler2019few} scenarios with proper instructions or prompts, even without parameter updates. However, OpenAI has not provided its training data so the question arises as to whether and to what extent ChatGPT can handle information extraction(IE) which is for applications like, question-answering and knowledge graph construction\cite{han2023information} evaluates ChatGPT's performance, evaluation criteria, robustness, and types of errors in information extraction tasks. By \cite{han2023information} ChatGPT is able to understand well the subject-object relationships in extraction tasks by fine tunnng it.  Large Language Models (LLMs) \cite{brown2020language} and prompt-based techniques such as PromptNER\cite{zhang2023promptner} is to use prompts or instructions to guide an LLM to recognize entities in a given sentence. This involves providing the LLM with a set of entity definitions and a few-shot example. The LLM\cite{brown2020language,thoppilan2022lamda} then generates a list of potential entities within the sentences in cross-domain NER PIXIU\cite{xie2023pixiu}.Additionally, a standardized benchmark covering financial NLP and prediction tasks that proposed to evaluate FinMA and existing LLMs.
The advancement of large language models (LLMs)\cite{brown2020language,thoppilan2022lamda} in natural language processing (NLP) within the financial domain is hindered by the absence of domain-specialized financial LLMs, instruction datasets, and evaluation benchmarks. Addressing this gap, PIXIU\cite{xie2023pixiu}, is a financial-based LLM based on the LLAMA framework features the financial LLM called FinMA\cite{xie2023pixiu}, which is achieved by refining LLaMA\cite{touvron2023llama} through instruction data fine-tuning that introduces a substantial instruction dataset of 136K data samples, including diverse financial tasks, document types, and data modalities. 
\subsection{Optical Character Recognition}
Optical Character Recognition (OCR) is a technology that converts printed or handwritten text from images, scanned documents, or photographs into machine-readable text. OCR involves several steps in its process. First, the input image is preprocessed, which includes tasks like noise reduction, binarization to convert it into black and white, and skew correction to align the text properly. Next, text regions are detected through methods like layout analysis and object detection, identifying areas containing text. The detected regions are then segmented into individual characters or words. The segmentation step is followed by feature extraction, where relevant characteristics of the characters are identified, aiding in recognition. The recognition phase involves matching extracted features with a predefined character set or language model to convert the image text into editable text. Finally, post-processing steps like spell-checking and grammar correction may be applied to refine the recognized text. The goal of OCR is to enable computers to understand and work with text data from images, enabling various applications such as document digitization, information retrieval, and text analysis.

OCR services provided by major cloud providers like Google Cloud\cite{cui2021document}, AWS \cite{bhatia2019comprehend}, and Azure\cite{kumar2023comparative} offer advanced capabilities for text extraction from images and documents. Google Cloud Vision OCR employs deep learning models to extract text, perform document structure analysis, and recognize handwriting. Azure Computer Vision OCR utilizes OCR technology combined with AI to extract printed and handwritten text, supporting multiple languages and document types. AWS Textract offers advanced features including text and form extraction, key-value pair identification, tabular data extraction, and its incorporation of a query answering system but it may require fine-tuning for optimal results. Despite the advantages of these services, limitations may include difficulties in handling handwritten text, complex layouts, and low-resolution images. It's important to consider specific requirements and data characteristics when choosing an OCR service.

Despite the significant advancements in Optical Character Recognition (OCR)\cite{cui2021document} technology, extracting complete key-value pairs from PDFs, particularly invoices and bills with diverse table layouts, remains a formidable challenge. The complexity arises from the varied and often unpredictable table formats within multi-structured invoice PDFs, which confound traditional OCR systems. Recognizing these table layouts accurately is an intricate task, and it is evident that there is a pressing need for more robust and adaptable approaches to address this issue. As to overcome these challenges by utilizing cutting-edge machine learning models, tailored OCR methods, and advanced table recognition algorithms. The primary objective is to facilitate accurate and thorough extraction of key-value pairs from these intricate document layouts, ultimately boosting data extraction efficiency, minimizing errors, and supporting the automation of vital processes. These insights underline the importance of continuous endeavors to enhance OCR capabilities, particularly in industries dependent on precise information retrieval from various complex document structures.

\section{Applications and its Challenges}

The application of Named Entity Recognition (NER) research spans numerous critical domains, each demanding tailored solutions to address distinct challenges and extract domain-specific knowledge effectively. In the realm of healthcare and biomedical research, NER is instrumental in unlocking insights from a burgeoning volume of medical data, enabling advancements in patient care, drug discovery, and disease monitoring. In the cybersecurity domain, NER aids in swiftly identifying and classifying cybersecurity entities, bolstering threat detection and incident response capabilities in the face of evolving cyber threats. Furthermore, NER empowers environmental scientists to decipher climate parameters, species data, and ecological trends, facilitating a deeper comprehension of environmental changes and conservation efforts. In the legal arena, NER streamlines the complex task of identifying legal entities, case references, and legal codes within vast legal documents, enhancing document analysis and compliance monitoring. Meanwhile, NER research in the energy sector supports informed energy policy development and sustainable resource management. During humanitarian crises, NER plays a pivotal role in swiftly locating and coordinating relief efforts for affected populations. In space exploration, NER catalogues celestial discoveries and space mission details, while in the AI and technology sector, it identifies emerging technologies and innovations, informing tech trend analysis and investment decisions. NER in education aids in research, student enrollment, and content recommendation, while in public health, it contributes to timely disease outbreak monitoring and resource allocation during health crises. Across these domains, NER research stands as a crucial and ever-evolving field, continuously adapting to diverse challenges and domains to deliver specialized entity recognition systems that enrich research, industry, and society as a whole.

\subsection{Finance}
\paragraph{Applications:}
Named Entity Recognition (NER) has several applications in the finance domain, mainly in the extraction of crucial information from various documents Applications of NER[4] in finance involve extracting important details from documents. For instance, it helps sort transactions by recognizing vendor names, product descriptions, and amounts. By linking a found entities to databases, NER provides extra info, like connecting a company's name to its financial data. It even helps analyze feelings by spotting entities in financial texts, making it possible to gauge opinions about companies, stocks, or events. Moreover, it's used to identify risk-related entities, aiding in evaluating potential financial risks. NER supports compliance with rules by pinpointing entities that matter for regulations. Also, it's handy in managing portfolios by recognizing stock symbols, company names, and financial markers.

\paragraph{Challenges:}
However, there are challenges when extracting named entity data from financial documents like PDFs, invoices, and bills. These documents come in different styles, making it hard to consistently find entities. Plus, some documents mix structured info (like tables) with unstructured text, needing special methods. Errors like typos or inconsistent naming in financial documents can affect accurate entity detection. Poor scans could lead to errors, and similar terms may mean different things based on context, needing smart NER models. Financial documents often have multiple entities together (like company names and money amounts), making it tricky to tell them apart. Also, there's not enough labeled data specific to finance, which makes building good models tough. Using specialized words and linking entities to external sources can be tough too, as can ensuring sensitive info is extracted in compliance with rules. Overcoming these issues needs smart NER models, special data, and methods that understand finance's complexities.

\subsection{Biomedical}

\paragraph{Applications:}

In the field of Biomedical, Named Entity Recognition (NER) has various uses. It helps in identifying crucial information in medical texts. For instance, it can locate and categorize medical terms like disease names, drug names, and medical procedures. NER also supports medical research by recognizing genes, proteins, and molecular structures mentioned in scientific literature. Moreover, it aids in tracking patient data, linking medical terms with patient records, and helps in managing medical databases.

\paragraph{Challenges:}
There are challenges in NER for Biomedical texts. One challenge is the vast number of medical terms and variations, which can make accurate recognition tricky. Also, medical texts often have complex sentence structures and abbreviations that can confuse NER systems. Ambiguities between common words and medical terms pose another issue. Additionally, there's a lack of large annotated datasets specific to Biomedical NER, making training accurate models challenging. Overcoming these hurdles requires advanced algorithms, specialized training data, and methods that can navigate the intricacies of medical language.

\subsection{Other applications and challenges}

Named Entity Recognition (NER) finds applications across various fields beyond finance and biomedical domains. In legal texts, NER helps identify legal terminologies, case references, and entities like names of laws and regulations. However, challenges arise due to the diverse legal language and context-specific entity meanings. In the news domain, NER aids in extracting people's names, locations, and organization names, supporting news categorization and sentiment analysis. The ambiguity of entity references and rapidly evolving entities pose challenges. Similarly, in the e-commerce sector, NER assists in extracting product names, brands, and specifications. Yet, the vast variety of product names and frequent changes in product listings create difficulties. In social media, NER is used for sentiment analysis, topic identification, and user profiling. However, the informal language, abbreviations, and context-dependent entity mentions make accurate recognition challenging. While Named Entity Recognition(NER) systems excel for English, numerous challenges persist for other Indian and Asian languages. The issues such as the absence of capitalization norms, linguist morphological complexities, the ambiguity between common and proper nouns, and the overlap of terms like person and location names. Overcoming these challenges requires domain-specific [12] training data, context-aware models, and adaptable algorithms.

\section{Conclusion}

In the realm of Natural Language Processing, Named Entity Recognition (NER) stands as a monumental pillar, bridging the gap between unstructured textual data and structured knowledge organization. This comprehensive survey of AI techniques for NER underpins the tremendous strides made in the domain, while also acknowledging the multifaceted challenges it continues to confront.

From the rudimentary rule-based methods to the transformative prowess of transformer architectures, the NER landscape has witnessed a profound metamorphosis. The dawn of domain-specific NER models accentuates the adaptability of these techniques, demonstrating the finesse with which models like ViBERTgrid and BioBERT cater to the nuanced requirements of specialized domains like finance and medicine. Yet, the continuous evolution of deep learning paradigms, like E-NER and the utilization of large language models, indicates that the NER journey is still unfolding, with newer horizons awaiting exploration.

The synergy between Optical Character Recognition (OCR) and NER is an exciting testament to the interdisciplinary collaboration within AI, showcasing the potential of integrating diverse technologies for more robust information extraction. Such amalgamations signal the future direction of NER — a direction characterized by holistic, multimodal information extraction that spans not only text but also visual and auditory data.

Despite these advancements, the challenges delineated in the paper underscore the areas awaiting deeper inquiry. The intrinsic hurdles posed by different industries, languages, and the ever-evolving nature of information, emphasize the dynamic terrain of NER. Adapting to low-resource languages and domains, while enhancing the model's ability to handle ambiguity and connect recognized entities to expansive databases, forms the crux of the road ahead.

Moreover, as NER cements its role in pivotal applications, from parsing intricate financial documents to deciphering the complexities of medical texts, its societal impact becomes even more pronounced. It becomes imperative to ensure the transparency, reliability, and ethical considerations of these models, considering their profound implications in decision-making processes across sectors.

In summation, while this research offers a panoramic view of the current NER landscape, it also serves as a clarion call to the research community. The realm of Named Entity Recognition is rife with opportunities, challenges, and responsibilities. With the accelerating pace of innovation and the ever-increasing significance of language understanding in the digital age, NER stands not just as a technical endeavor but as a cornerstone in shaping the future of information extraction and knowledge representation. The road ahead is long, promising, and teeming with potential — a journey that beckons the collaborative efforts of researchers, practitioners, and industry stalwarts.

\bibliographystyle{unsrt}  
\bibliography{references}
\end{document}